\documentclass[conference]{./sty/IEEEtran}

\usepackage[OT1]{fontenc} 

\IEEEoverridecommandlockouts
\usepackage{cite}
\usepackage{amsmath,amssymb,amsfonts}
\usepackage{graphicx}
\usepackage{textcomp}
\usepackage{xcolor}
\usepackage{hyperref}
\usepackage{bm}
\usepackage{soul}
\graphicspath{{pic/}}
\usepackage{xspace}

\newcommand{\etal}{\textit{et al}.}
\newcommand{\figref}[1]{Fig. \ref{#1}}

\def\BibTeX{{\rm B\kern-.05em{\sc i\kern-.025em b}\kern-.08em
    T\kern-.1667em\lower.7ex\hbox{E}\kern-.125emX}}
\begin{document}

\title{Web-based Experiment on Human Performance in Dual-Robot Teleoperation
}


\author{\IEEEauthorblockN{Yuhui Wan and Chengxu Zhou}
\IEEEauthorblockA{\textit{School of Mechanical Engineering} \\
\textit{University of Leeds}\\
Leeds, United Kingdom \\
\{mnywa, c.x.zhou\}@leeds.ac.uk}
}

\let\oldmaketitle\maketitle
\renewcommand{\maketitle}{\oldmaketitle\setcounter{footnote}{0}}

\maketitle

\begin{abstract}
In most cases, upgrading from a single-robot system to a multi-robot system comes with increases in system payload and task performance. On the other hand, many multi-robot systems in open environments still rely on teleoperation. Therefore, human performance can be the bottleneck in a teleoperated multi-robot system. Based on this idea, the multi-robot system's shared autonomy and control methods are emerging research areas in open environment robot operations. However, the question remains: how much does the bottleneck of the human agent impact the system performance in a multi-robot system? 
This research tries to explore the question through the performance comparison of teleoperating a single-robot system and a dual-robot system in a box-pushing task. This robot teleoperation experiment on human agents employs a web-based environment to simulate the robots' two-dimensional movement. The result provides evidence of the hardship for a single human when teleoperating with more than one robot, which indicates the necessity of shared autonomy in multi-robot systems.

\end{abstract}

\section{Introduction}
 Many tasks still require human involvement, especially in outdoor environments where structures and objects can be unpredictable. Human living space is an excellent example of this environment, where both mission and surroundings have uncertainty. Also, these tasks may require multiple robots to complete. For example, moving a table requires more than one mobile manipulator. The multi-robot teleoperation simultaneously creates a heavy workload on the human operator \cite{Human}. Finding an efficient way for a single operator to control one or more robots has been an emerging area. The foundation of developing such a system is understanding human limitations and the interaction between humans and robots.
 
 Choosing an efficient interface for these users is critical, especially when a fully automated system is impossible. Over the years, studies have been carried out to develop efficient shared autonomy systems \cite{fontaine2021evaluating}, and even for multi-robot systems \cite{LIANG2021468}. However, most research mainly focuses on the robots and hardware side of shared autonomy. Furthermore, limited research evaluates the performance of human agents and human-robot teleoperation interfaces (HRTIs) \cite{Wan}.
 
 Experiments involving human interaction with a robot are expensive and time-consuming, even in a simulation, due to the system and hardware requirements for robot simulation software, including ROS and Matlab. Therefore, human participants still need to use the designated computer physically. A web-based system can be more efficient by allowing participants to perform experiments remotely through the internet. Open Graphics Library (OpenGL), a web-based interface, gives the browser access to the computer's GPU. This makes GPU-based 2D and 3D environments available to the browser. Based on this ideal, Web Graphics Library (WebGL) has become one of the most popular forms of web-based simulation and visualisation form, which can be applied to the robot visualisation area \cite{WebGL}. 
 
 In this work, we deploy a WebGL-based system to compare the performance of a single operator teleoperating one or two robot agents simultaneously to complete a box-pushing mission. Although a dual-robot system is harder to control, two robots provide double the power and can move payload faster. The experiment allows participants to take part remotely by browsing a website. Furthermore, we collect the movement time to complete the same mission using a single-robot and a dual-robot system.

\begin{figure}
    \centering
    \includegraphics[width=0.9\columnwidth]{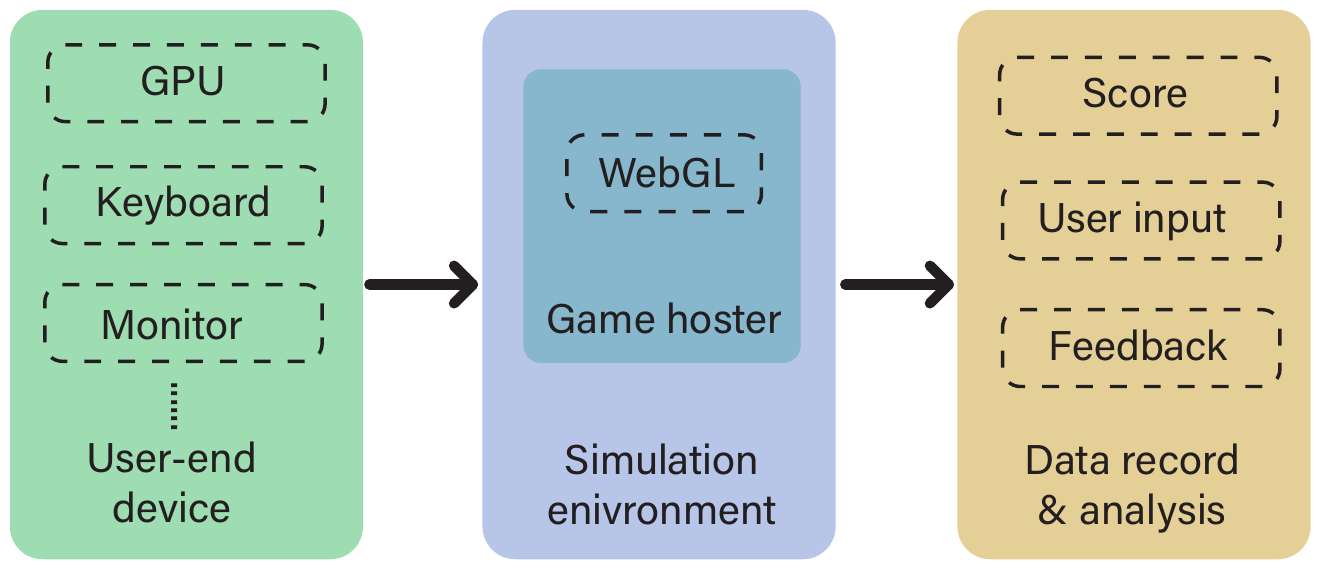}
    \vspace{-2mm}
    \caption{The structure of the web-based system for HRTI simulation.}
    \label{fig:Structure}
\end{figure}

\section{System Overview}
As illustrated in \figref{fig:Structure}, the developed simulation system comprises three parts, the user-end device, the WebGL-based online simulation environment, and the data analysis.

In this research, we designed an experiment to discover how humans perform in the teleoperation of a single robot\footnote{\url{https://play.unity.com/mg/other/single-robot-teleoperation}} and a dual-robot\footnote{\url{https://play.unity.com/mg/other/dual-robot-teleoperation}} system. Participants take part in the experiment remotely through a web-based simulation game, where the simulation environment runs on user-end device hardware.

\subsection{User-end devices}
WebGL can access the device's GPU from the browser through HTML5 elements without additional plug-ins. This makes it possible for web-based systems to take advantage of high-performance GPUs in the user's device. Therefore, the requirement for user-end devices varies based on the requirements of the visual environment.

In this research, the targeted environment is in two dimensions and has minimal system hardware requirements. However, a keyboard and monitor are still required as human-machine interference. 

\subsection{WebGL based environment and experiment}
One of the most popular WebGL developing environments is ``Unity''. The experiment in this study benefits from the Unity resource library and free online hosting server ``Unity Play''. Therefore, a simulation game is developed with more attractive scenery and is available to any potential participants. Also, the delay due to the internet is eliminated due to WebGL performing computation on participants' local machines.

\begin{figure}
    \centering
    \includegraphics[width=0.85\columnwidth]{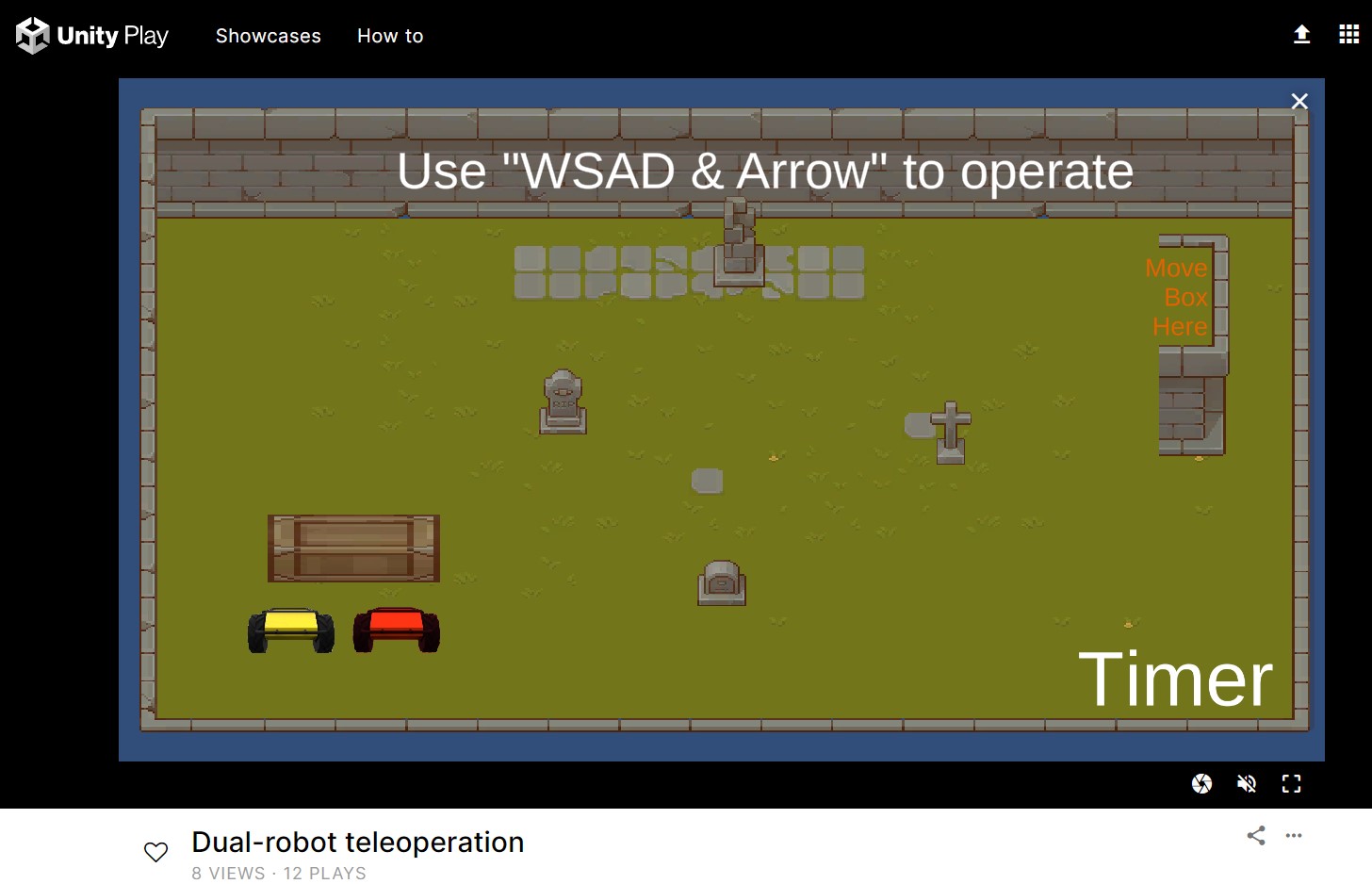}
    \vspace{-3mm}
    \caption{Example WebGL game hosted by Unity Play. Two robots in red and yellow start below the targeted box at the lower left of the screen. The finish area is marked in ``Move Box Here'' at the upper right of the screen.}
    \label{fig:game}
\end{figure}

In the real world, upgrading from a single-robot to a multi-robot system can increase the total power and complete tasks faster. This research compares the performance of a single-robot and a dual-robot system with the same robot agents in the same teleoperation task. In other words, whether one additional robot can directly increase system performance. In this mission, the robot agent(s) push a box from the starting point to the finish area, as shown in \figref{fig:game}, and movement time is measured for performance. Although the experiment environment is in 2D, there are still rigid bodies and dynamic models for each object. In the physical model of simulation, each robot can provide the same amount of power. In the setting, two robots together can push the box 50\% faster than a single robot. There are four additional obstacles between the start and the finish area to increase task difficulty.

\subsection{Data analysis}
A server could be employed to help collect recorded data, including users' scores, robot trajectory history, and feedback. However, this research only collects and analyses the movement time to complete the task, which can be traded as scores, where the lower, the better.

\section{Experiment Performance and Result}
\begin{figure}
    \centering
    \includegraphics[width=\columnwidth]{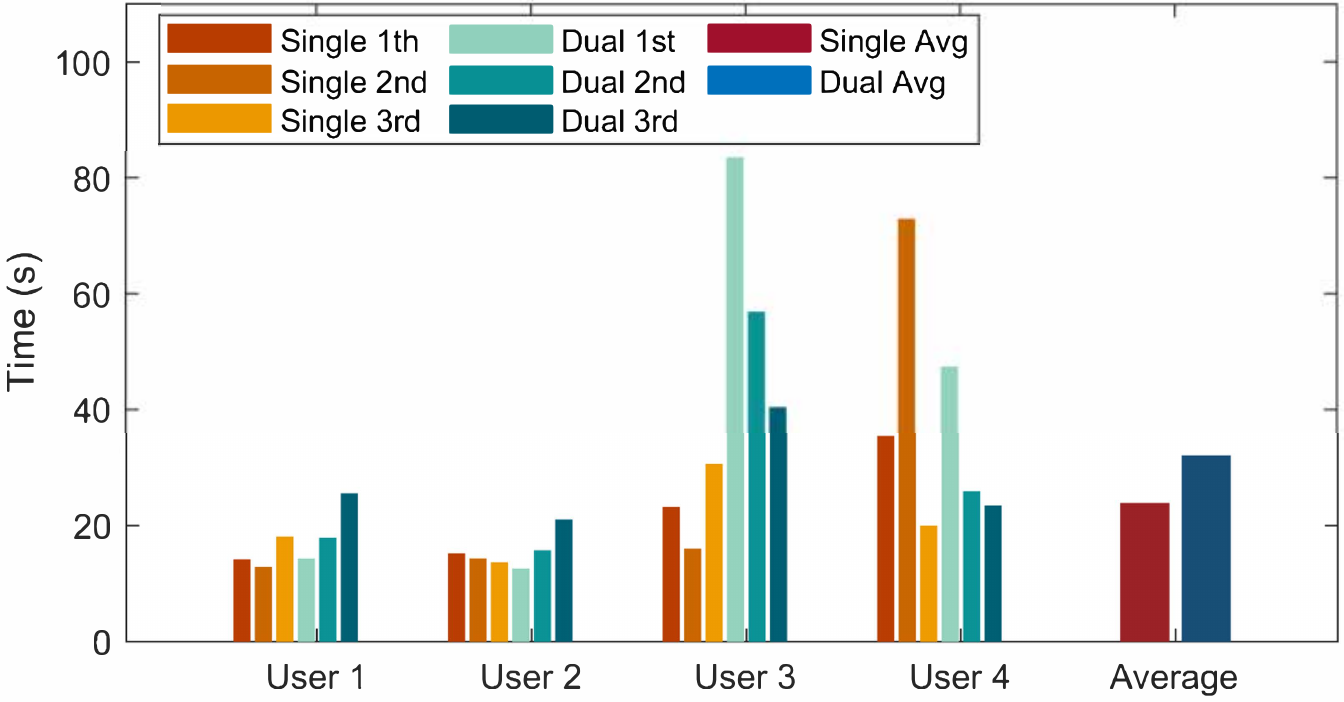}
    \vspace{-5mm}
    \caption{Movement time of each user's trials with both single-robot system and dual-robot system, and the average time among all users.}
    \label{fig:Big}
\end{figure}

Four participants took part in the experiment remotely with the web-based simulation environment. The study measures their first three attempts' movement time, from any key pressed until the box reaches the finish area. In addition, they experimented with random orders. For example, users 1 and 4 performed the task with a single robot and then with a dual robot, while others performed the task in reverse order. The movement time for each user's attempts is shown in \figref{fig:Big} with the average time for two different robot groups. 

In theory, the dual-robot system with more dynamic power can push the box faster, and two robots have an advantage in balancing the box in moving. However, the average movement time is similar, with the double-robot system taking 34.3\% more time to complete the task than the single-robot system. Therefore, for the same task, simply adding a robot to a teleoperating system can not increase the system performance without an additional control method and a better HRTI. 


\section{Conclusion}\label{SCM}
The research used a web-based simulation environment to study the performance of human-robot interaction. The performance difference of a single user teleoperating one and two robots simultaneously to complete a simple mission. The experiment result shows the difficulty of humans teleoperating with two robot agents simultaneously, even though the system is in only two dimensions. 

The study have explored the limits of a single human agent interacting with multiple robots. Further research will target this limitation of human agents and design a shared autonomy system for better multi-robot control.


\end{document}